\documentclass{article}
\pdfoutput=1

\usepackage{times}
\usepackage{graphicx}
\usepackage{subfig}

\usepackage{algorithm}
\usepackage{algorithmic}
\usepackage{hyperref}

\usepackage{amsmath,amsfonts}
\usepackage{commath}
\usepackage{verbatim}
\usepackage{bbm}
\usepackage{multirow}

\DeclareMathOperator*{\argmin}{arg\,min}
\DeclareMathOperator*{\argmax}{arg\,max}

\begin{document}

\newcommand{\method}{VRD}
\renewcommand{\dif}{\mathop{}\!\mathrm{d}}
\newcommand*{\tran}{^{\mkern-1.5mu\mathsf{T}}}
\newcommand{\so}{\ensuremath{s^o}}
\newcommand{\tso}{\ensuremath{\tilde s^o}}
\newcommand{\si}{\ensuremath{s^i}}
\newcommand{\spr}{\ensuremath{s^p}}
\newcommand{\No}{\ensuremath{{N_o}}}
\newcommand{\Ni}{\ensuremath{{N_i}}}
\newcommand{\Bo}{\ensuremath{B^o}}
\newcommand{\Bi}{\ensuremath{B^i}}
\newcommand{\bBo}{\ensuremath{\bar{B}^o}}
\newcommand{\Qo}{\ensuremath{Q^o}}
\newcommand{\bQo}{\ensuremath{\bar{Q}^o}}
\newcommand{\Qi}{\ensuremath{Q^i}}

\title{Variational reaction-diffusion systems for semantic segmentation}

\author{Paul Vernaza\\NEC Laboratories America\\10080 N. Wolfe Road, Cupertino, CA 95014\\\texttt{<pvernaza@nec-labs.com>}}

\maketitle

\begin{abstract} 
  A novel global energy model for multi-class semantic image
  segmentation is proposed that admits very efficient exact inference
  and derivative calculations for learning.
  Inference in this model
  is equivalent to MAP inference in a particular kind of vector-valued
  Gaussian Markov random field, and ultimately reduces to solving a
  linear system of linear PDEs known as a reaction-diffusion system.
  Solving this system can be achieved in time scaling near-linearly
  in the number of image pixels by reducing it to sequential FFTs,
  after a linear change of basis.
  The efficiency and differentiability
  of the model make it especially well-suited for integration with
  convolutional neural networks, even allowing it to be used 
  in interior, feature-generating layers and stacked multiple times.
  Experimental results are shown demonstrating that the model
  can be employed profitably in conjunction with different convolutional
  net architectures, and that doing so compares favorably to 
  joint training of a fully-connected CRF with a convolutional net.
\end{abstract} 

\section{Introduction}

The focus of this work is the semantic segmentation problem, in which
a learning system must be trained to predict a semantic label for each
pixel of an input image.  Recent advances in deep convolutional neural
nets (CNNs), along with historical successes of global energy methods
such as Conditional Random Fields (CRFs), have raised the natural
question of how these methods might best be combined to achieve better
results on difficult semantic segmentation problems.  Although several
proposals have recently emerged in this
vein~\cite{chen14semantic,zheng2015conditional,schwing2015fully,lin2015Piecewise},
there is currently no clear consensus on how best to integrate these
methods.

\begin{figure}
  \centering      
  \includegraphics[width=3.25in]{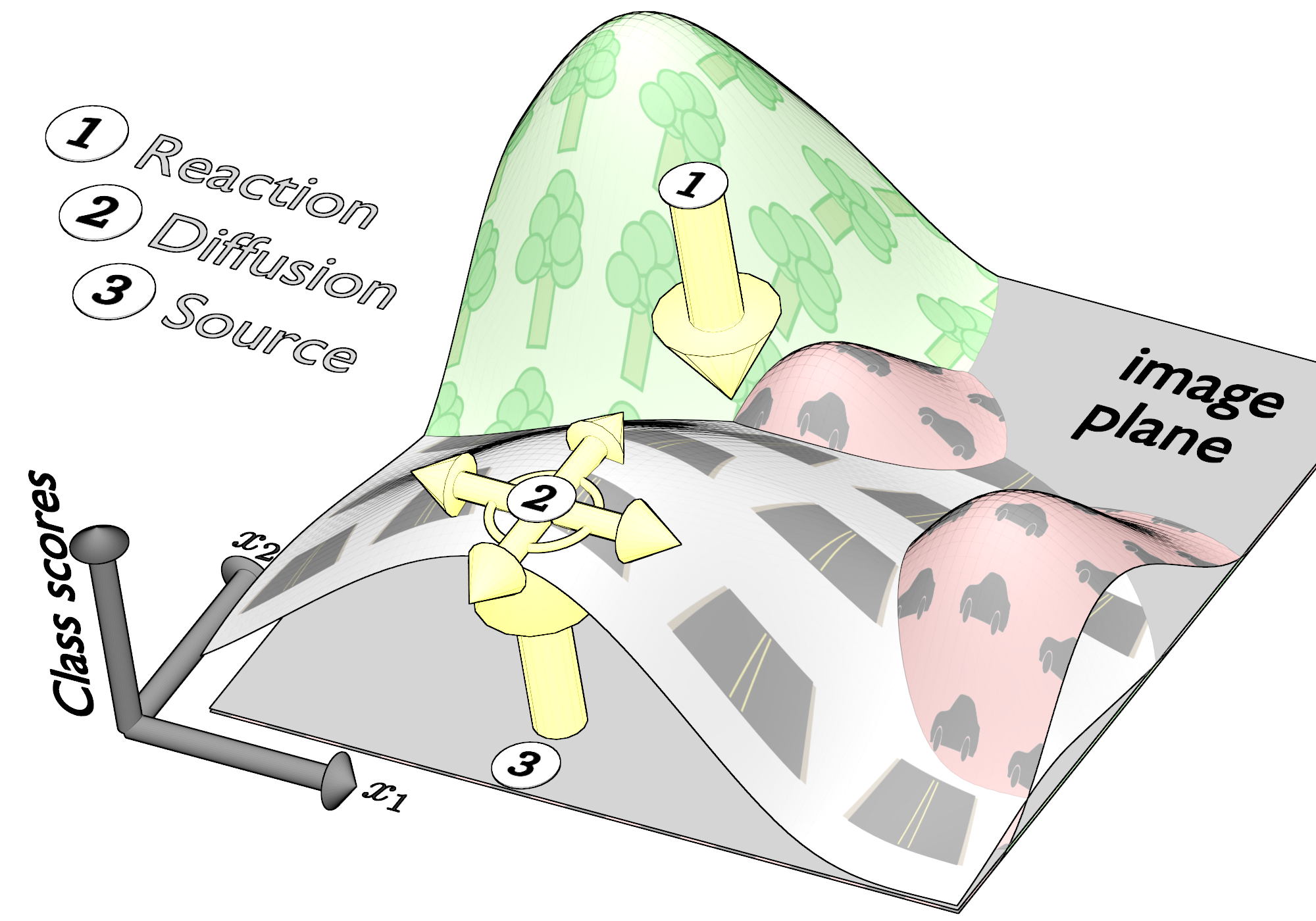}
  \caption{Illustration of reaction-diffusion analogy for an image
    segmentation task with three classes: road, tree, and car. }
  \label{fig:rd}
\end{figure}

Achieving tighter integration between and better joint training of
global energy models and CNNs is the key motivator for this work.  To
that end, this paper proposes a novel global energy model for semantic
segmentation, referred to here as Variational Reaction Diffusion (or
\method{}).  \method{} can be thought of as a vector-valued Gaussian
Markov Random Field (GMRF) model over a continuous domain (as opposed to 
a graph).  Unlike most other energy-based methods, {\em exact}
inference in \method{} can be performed very efficiently by reducing
the problem to sequential FFTs, after a linear change of basis.
Backpropagation and parameter derivatives can also be computed
efficiently, making it an attractive choice for integration with CNNs.
The efficiency of \method{} also raises some interesting new
possibilities for tight integration of CNNs with global energy
methods: instead of appending a relatively complex CRF model to an
existing CNN architecture, \method{} may actually be used as an
internal, feature-generating layer, for example. This possibility is
explored in the experiments.

Since inference in \method{} is linear in the inputs, 
an obvious concern is whether such a simple model manages 
to capture the most important features of seemingly more complex
models requiring approximate inference or sophisticated combinatorial
optimization.  Although the possibility of layering somewhat negates
this concern, Section~\ref{sec:analysis} also provides some insight
into this issue by showing how \method{} can be considered a
relaxation of other popular models.  Experiments in
Section~\ref{sec:experiments} also shed some light on this question by
showing that \method{} compares favorably to more complex energy-based
methods.

The name of the proposed method is a reference to the
reaction-diffusion systems that initially inspired this work.
Briefly, inference in \method{} may be interpreted as evolving
evidence (or class scores) under the dynamics of a reaction-diffusion
process, as illustrated in Fig.~\ref{fig:rd}. Intuitively, we might
think of modeling evidence for one semantic class as being created by
unary potentials (or the previous layer in a CNN), propagating across
the image via diffusion, and reacting with evidence for other semantic
classes. Each of these processes may locally create or suppress
evidence for each class, and if we allow this process to reach an
equilibrium, the sum of these effects must cancel at every point in
the image (c.f. Eq.~\ref{eq:rdSysPDE}).  By restricting the model to
the class of such processes generating the solutions to convex,
variational problems, we are essentially ensuring that such an
equilibrium exists and is globally stable.

The rest of this paper is structured as follows.  The next section
gives a very brief overview of the method and a summary of the results
that make inference and learning tractable.
Section~\ref{sec:analysis} motivates the model by comparing it to
existing models, gives some intuition as to how inference works, and
discusses other practical issues.  The main results for inference
and learning \method{} are derived in Section~\ref{sec:derivation}.
This is followed by a discussion of related work and experimental results.

\section{Method overview} \label{sec:overview}

This section gives a brief overview of the main ideas and results of the method.
Details will be discussed subsequently.

\subsection{The model}

Let $I \subset \mathbb R^2$ denote the image plane: i.e., a
rectangular subset of $\mathbb R^2$ representing the domain of the
image.  \method{} is given a spatially-varying set of $\Ni$ input
features, represented here as a function
$\si : I \rightarrow \mathbb R^\Ni$,
and produces a set of $\No$ output scores
$\so : I \rightarrow \mathbb R^\No$.
For now, $\No$ might be thought of as the number of
semantic classes, and we might think of $\so_k(x)$
as a score associated with the $k$th class at
$x \in I$, with a prediction generated via $\argmax_k \so_k(x)$.
Throughout this paper, $x$ will represent an arbitrary
point in $I$.

Let $s = \begin{pmatrix} \so{}\tran & \si{}\tran \end{pmatrix}\tran$
denote the concatenation of $\si$ and $\so$ into a single
function $I \rightarrow \mathbb R^{\Ni + \No}$.
\method{} generates $\so$ by solving the following optimization problem.
In the following, the dependence of $s$ on $x$ has been omitted for clarity.
\begin{equation}
  \argmin_{\so}
  \int_I
  s\tran Q s
  + \sum_{k=1}^2 \dpd{s}{x_k}\tran B \dpd{s}{x_k}
  \dif x.\label{eq:obj}
\end{equation}
Here, $B$ and $Q$ are assumed to be constant (i.e., independent of 
$x$) positive-definite parameter matrices.  This is then an
infinite-dimensional, convex, quadratic optimization problem in $\so$.

\subsection{Inference}

Just as the minimum of a convex, finite-dimensional quadratic function
can be expressed as the solution to a linear system, the solution to
this infinite-dimensional quadratic can be expressed as the solution
to the following linear system of PDEs:
\begin{equation}
  \Bo \Delta \so - \Qo \so = \Qi \si - \Bi \Delta \si,\label{eq:rdSysPDE}
\end{equation}
where the dependence on $x$ has again been omitted, $\Delta$
represents the vector Laplacian ($(\Delta f)_i := \sum_j
\pd[2]{f_i}{x_j}$), and $B$ and $Q$ have been partitioned into
submatrices \Bo, \Qo, \Bi, and \Qi{} such that
$s\tran Qs = \so{}\tran \Qo \so + 2 \so{}\tran \Qi \si + f(\si)$
 (and likewise for $B$).  We can solve this
system efficiently via a linear change of variables and a
backsubstitution procedure exactly analogous to the finite-dimensional
case.  Specifically, we first use the Schur decomposition to write
$(\Bo)^{-1} \Qo = VUV\tran$, where $V$ is orthonormal and $U$ is
upper-triangular.  We then perform the change of variables $z =
V\tran\so$.  Let $\spr := \Qi \si - \Bi \Delta \si$.  We then solve for
$z$ via backsubstitution, first solving the following scalar PDE for
$z_\No$, fixing it, solving for $z_{\No-1}$, and proceeding thus backwards to $z_1$:
\begin{equation}
  \Delta z_k - U_{kk} z_k = (V\tran(\Bo)^{-1} \spr)_k + \sum_{j=k+1}^\No U_{kj} z_j.\label{eq:backsub}
\end{equation}
After solving for $z$, the output scores are obtained via $\so = V z$.
The scalar PDEs above may be discretized and solved either via the FFT
or the multigrid method~\cite{brandt1982guide}. If $L$ lattice points
are used in the discretization, the total computational cost of
solving~\eqref{eq:obj} via FFT-based inference is $O(\No^3 + \No^2 L +
\No L \log L + \No \Ni L)$.

\subsection{Learning} \label{sec:overviewLearning}

In order to learn the model, we assume some arbitrary, differentiable loss
$L(\so)$ has been defined on the output scores \so{}.
Gradient-based learning is enabled by computing the derivatives
of $L$ with respect to the parameter matrices $B$, $Q$, and
potentially the inputs \si{}, allowing the model to be used in
backpropagation.

The backpropagation derivative $\od{L}{\spr{}} : I \rightarrow \mathbb R^\No$
(with \spr{} defined as above) can be computed by solving the same PDE
system~\eqref{eq:rdSysPDE} as in the inference step, but replacing
$\spr$ with $\od{L}{\so{}}$.  Specifically, we solve
\begin{equation}
  \Bo \Delta \dod{L}{\spr{}} - \Qo \dod{L}{\spr{}} = \dod{L}{\so{}} \label{eq:dLdsp}
\end{equation}
for $\od{L}{\spr{}}$, given $\od{L}{\so{}} : I \rightarrow \mathbb R^\No$,
in the same way as in the inference step.
The parameter derivatives can be expressed as simple functions
of the backpropagation derivative.  These are as follows:
\begin{align}
  \dod{L}{{\Bo_{ij}}} &=
  - \left\langle \dod{L}{{\spr_i}}, \Delta \so_j \right\rangle \label{eq:dLdB}\\
  \dod{L}{{\Qo_{ij}}} &=
  \left\langle \dod{L}{{\spr_i}}, \so_j \right\rangle, \label{eq:dLdQ}
\end{align}
where the inner product is defined in the standard way, as $\langle f,
g \rangle := \int_I f(x) g(x) \dif x$, and $\so$ is that computed
via inference for the current values of $B$ and $Q$.

\section{Analysis} \label{sec:analysis}

\subsection{Comparison to other energy-based models}

Although the model~\eqref{eq:obj} may appear quite different from
other energy-based models due to the continuous formulation, the
motivation for and structure of the model is very similar to that for
more familiar models.  A typical CRF model for segmentation represents
a distribution over (class-label-valued) functions defined on the
nodes of a graph, which represent pixels or regions.  Inference using
the mean-field approximation reduces to finding a simplex-valued
function on the graph (representing the pseudo-marginal label distributions),
subject to some local self-consistency conditions.  Unary potentials
serve to anchor the distributions at each point, while binary potentials
encourage smoothness of these distributions with respect to neighboring points.

\begin{figure}
  \centering
  \includegraphics[width=2.25in]{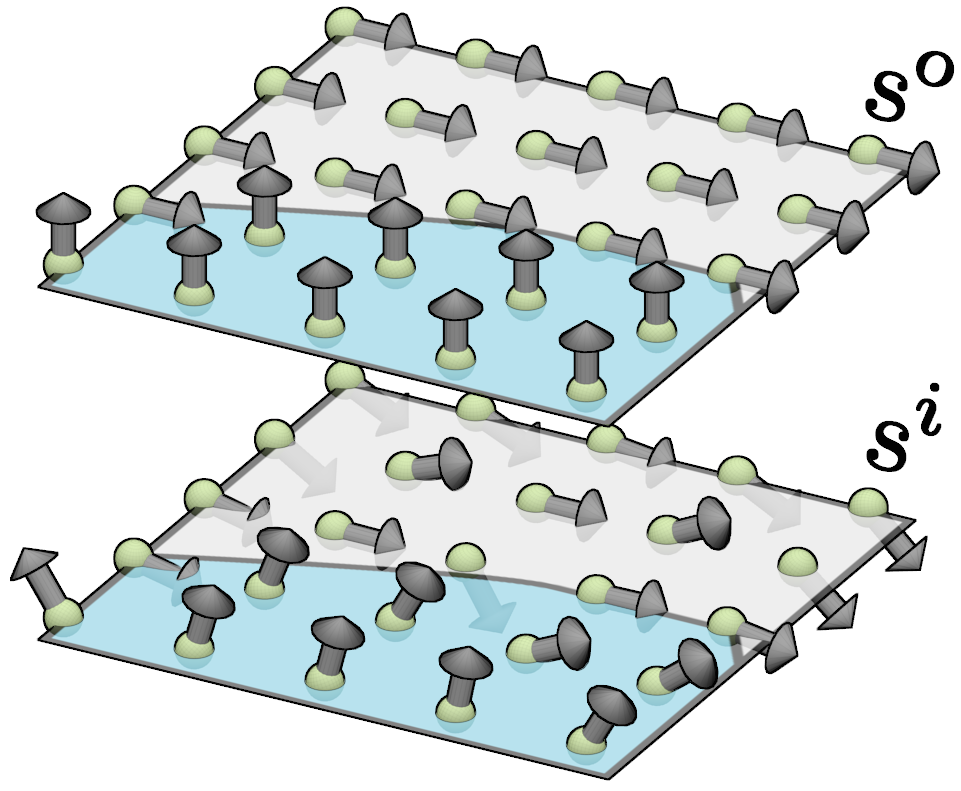}
  \caption{Illustration of variational model. Fig.~\ref{fig:vecField}
    depicts the discretized model: at each grid location (green ball),
    we are given an input vector (arrows on lower plane). For
    appropriate choice of $Q$, input vectors $s^i$ can be thought of
    as noisy estimates of output vectors $s^o$, which are to be
    inferred. Blue-shaded area represents one class, while white area
    represents other class.  Learning essentially trains output arrows
    in blue area to point upwards and arrows in white area to point to
    the right.}
  \label{fig:vecField}
\end{figure}

\begin{figure}
  \centering
  \subfloat[]{
    \includegraphics[width=1.5in]{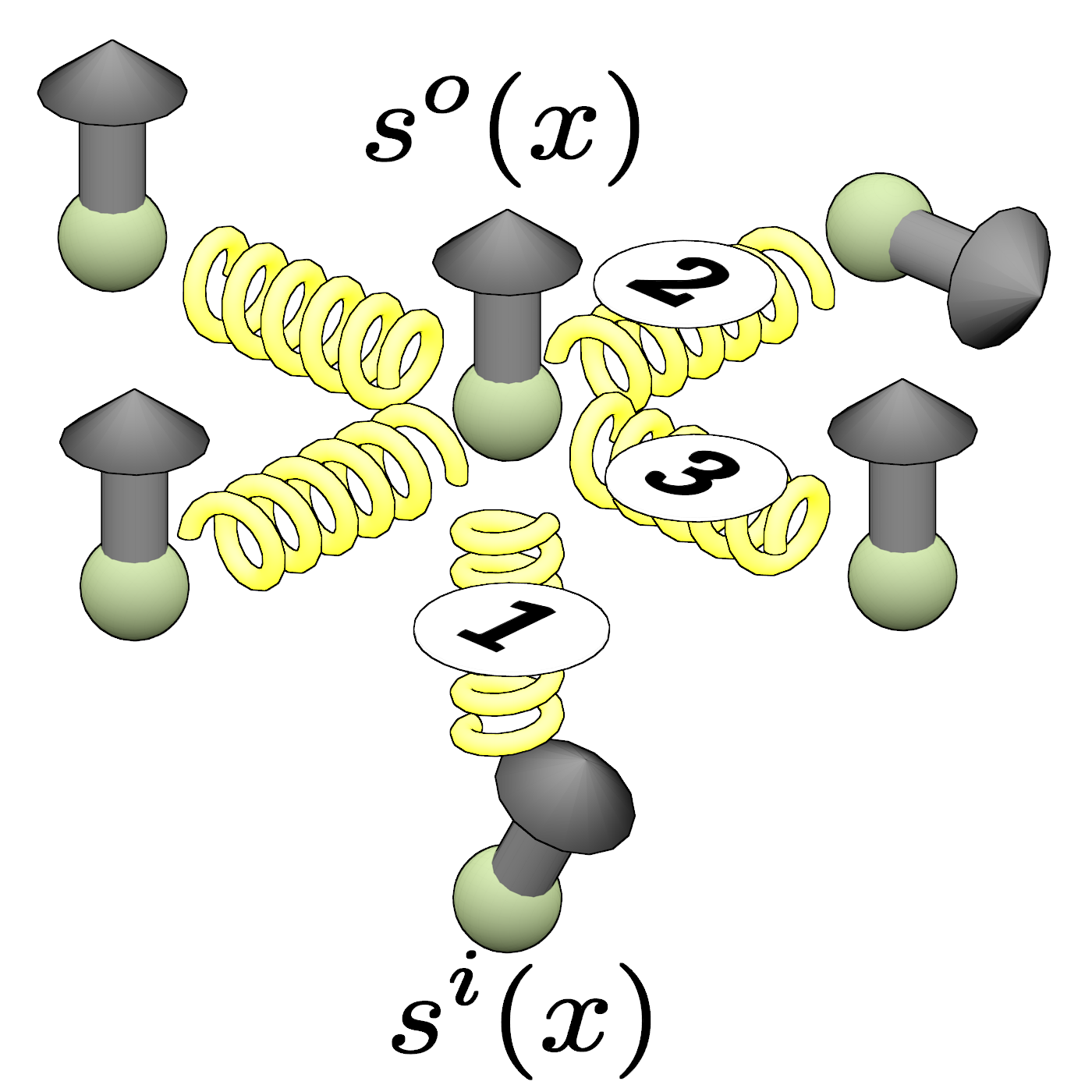}\label{fig:springs}
  }
  \subfloat[]{
    \includegraphics[width=1.5in]{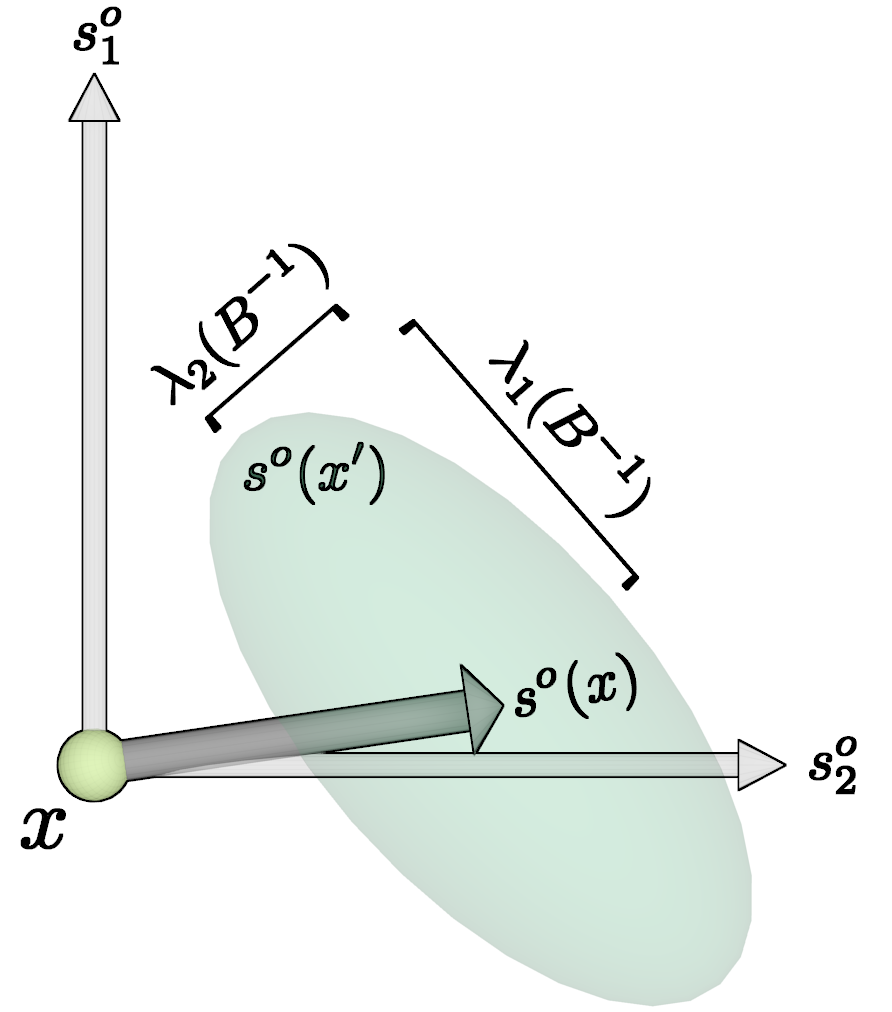} \label{fig:edgePot}
  }
  \caption{Fig.~\ref{fig:springs}: depiction of quadratic unary and
    binary potentials. Spring marked (1) corresponds to unary term
    $s\tran Q s$, spring marked (2) corresponds to binary term
    $\pd{s}{x_2}\tran B \pd{s}{x_2}$, and spring marked (3)
    corresponds to binary term $\pd{s}{x_1}\tran B \pd{s}{x_1}$.
    Fig.~\ref{fig:edgePot}: illustration of (log) Gaussian form of binary
    potential. For each location $x$, the potential is high when
    neighboring vector $\so(x')$ is outside an ellipse centered at
    $\so(x)$ with axes determined by $B$.}
\end{figure}

By contrast, \method{} produces a vector-valued score function defined
on $I$.  The derivatives in~\eqref{eq:obj} can be replaced by
finite-difference approximations to obtain a model defined over a
regular lattice graph.  This is illustrated in
Fig.~\ref{fig:vecField}.  Unary and binary potentials are both
quadratic, as illustrated in Fig.~\ref{fig:springs}.  Since all the
potentials are quadratic, the overall energy of~\eqref{eq:obj} can be
thought of as the unnormalized log-likelihood of a Gaussian Markov random
field, albeit a vector-valued variant.

It is therefore evident that the principal difference between the
mean-field CRF and \method{} models is that \method{} relaxes the
probability simplex constraints on its outputs, and instead produces
unnormalized score functions.  \method{} also assumes the additional structure
that the domain of the modeled function must be equivalent to
Euclidean space, and the unary and binary potentials must be
quadratic.  These key assumptions enable very efficient, exact inference.

\subsection{Motivation for binary potential}

The quadratic binary potential in~\eqref{eq:obj} can be thought of as
a natural extension of the standard Potts model commonly employed
in energy-based methods.  To make this clear, consider a
finite-difference approximation of the quadratic term in~\eqref{eq:obj}.
Denoting by $\delta_k$ a unit vector aligned to axis $k$, we have
\begin{equation}
  \dpd{\so{}}{x_k}\tran B \dpd{\so{}}{x_k} \approx
  \epsilon^{-2} \Vert \so(x+\epsilon \delta_k) - \so(x) \Vert_B^2,
\end{equation}
where $\epsilon$ is a small step size.  If $\so$ were a binary
indicator vector ($\so_j(x) = 1 \iff \textrm{label}(x) = j$), and we
had $B = I$, then this term would correspond exactly to the Potts potential
$\mathbbm 1\{ \textrm{label}(x) \neq \textrm{label}(x + \epsilon \delta_k) \}$.
Fig.~\ref{fig:edgePot} illustrates the effect of the binary
potential for the general case: it essentially serves as a Gaussian
prior on the difference between score vectors of neighboring points.

\subsection{Comparison with submodular combinatorial optimization}

Here it is shown that the assumption of convexity of~\eqref{eq:obj} is
akin to the common assumption of submodular potentials in combinatorial 
global energy models.  In particular, it is demonstrated that in the
binary-label case, a discretized special case of~\eqref{eq:obj} corresponds
to a continuous relaxation of a binary, submodular optimization
problem.

Fixing $\Qo = I$, $\Qi = -I$, $\Bi = 0$, discretizing~\eqref{eq:obj}
via finite differences, and defining an appropriate lattice graph
with $\epsilon$-spaced nodes $\mathcal N$ and edges $\mathcal E$ yields
\begin{align}
  \argmin_{\so} \sum_{x \in \mathcal N} &\Vert \so(x) \Vert^2 - 2 \so(x)\tran \si(x) \notag\\
  &+ \sum_{(x,x') \in \mathcal E} \epsilon^{-2} \Vert \so(x') - \so(x) \Vert_{\Bo}^2.\label{eq:objFD}
\end{align}
An analogous combinatorial optimization can be defined by optimizing over
binary indicator vectors instead of \so{}.  Let $\mathbf 1_j \in \{0,1\}^\No$
denote the vector that is 1 in the $j$th position and 0 elsewhere. 
The analogous combinatorial optimization is then
\begin{align}
  \argmin_{l} \sum_{x \in \mathcal N} &\Vert \mathbf 1_{l(x)} \Vert^2 - 2 \mathbf 1_{l(x)}\tran \si(x)\\\notag
  &+ \sum_{(x,x') \in \mathcal E} \epsilon^{-2} \Vert \mathbf 1_{l(x')} - \mathbf 1_{l(x)} \Vert_{\Bo}^2.
\end{align}
The term $E_b(i,j) := \Vert \mathbf 1_i - \mathbf 1_j \Vert_{\Bo}^2$ is
referred to as the binary potential.
In the binary-label case ($\No = 2$), this optimization is said to be submodular
if the following condition holds:
\begin{equation}
  E_b(0,0) + E_b(1,1) \leq E_b(0,1) + E_b(1,0).
\end{equation}
In our case, we have $E_b(0,0) = E_b(1,1) = 0$ and
$E_b(1,0) = E_b(0,1) = \Bo_{00} + \Bo_{11} - \Bo_{10} - \Bo_{01}$,
which is nonnegative by the convexity assumption, since convexity 
requires that $\Bo$ be positive semidefinite.
This implies that the combinatorial analog of~\eqref{eq:objFD}
is submodular in the binary-label case.  Equivalently, we
may interpret~\eqref{eq:objFD} as a relaxation of a combinatorial
optimiziation obtained by relaxing the integrality and
simplex constraints on \so{}.  This may also be compared to
LP relaxation, which would relax the integrality constraint, but
retain the simplex constraints, at the expense of harder optimization.

\subsection{Intuition for inference}

\begin{figure}
  \centering
  \subfloat[$U_{kk} = 10^{-2}$]{
    \includegraphics[width=1.5in]{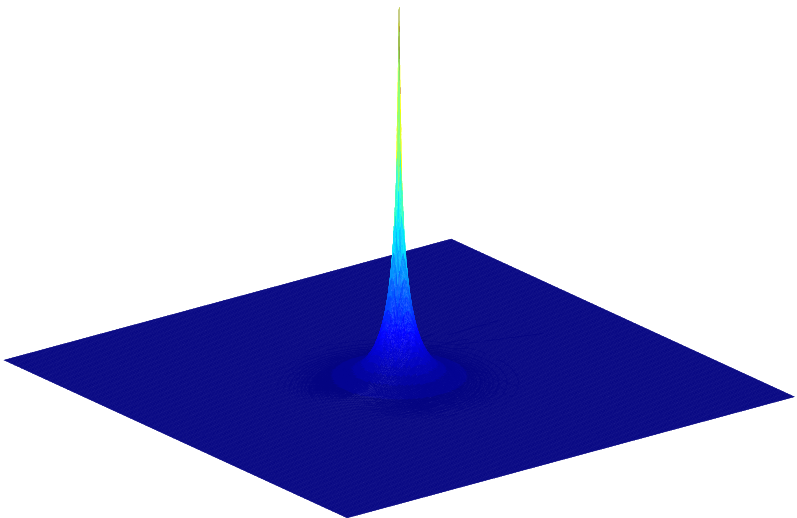}
  }
  \subfloat[$U_{kk} = 10^{-6}$]{
    \includegraphics[width=1.5in]{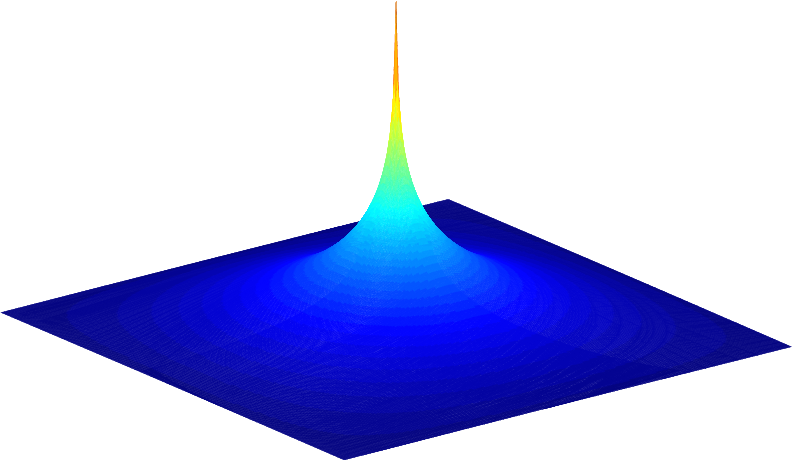}
  }
  \caption{The Green's function for the scalar PDE~\eqref{eq:backsub},
    for varying values of $U_{kk}$. The PDE can be solved by convolving the
    right-hand side with this function.}
  \label{fig:greens}
\end{figure}

The key step in performing inference is the solution of the scalar
PDEs~\eqref{eq:backsub}.  In the case of constant $B$ and $Q$, this
can be solved by taking the Fourier transform of both sides, solving
algebraically for the transform of the solution, and then inverting
the transform.  This process is equivalent to convolving the
right-hand side in the spatial domain with the {\em Green's function},
which is illustrated in Fig.~\ref{fig:greens}.  It is therefore
evident that for large values of $U_{kk}$, solving~\eqref{eq:backsub}
essentially convolves the right-hand side with a delta function (i.e.,
applying the identity function), while the solution for small values
of $U_{kk}$ convolves the right-hand side with an edge-preserving
filter with a very wide support.  Recalling that the $U_{kk}$ are the
(positive) eigenvalues of $(\Bo)^{-1} \Qo$, it is also therefore evident
that the amount of smoothing scales with the scale of $\Bo$ and
inversely with the scale of $\Qo$.  Intuitively, this means that the
smoothing decreases as the unary penalty grows and increases as the
binary penalty grows, just as one might expect.

In practice,~\eqref{eq:backsub} is solved via discretization and
discrete Fourier transforms.  Specifically, the right-hand side
of~\eqref{eq:backsub} is discretized over a regular grid, and the goal
is then to obtain samples of $z_{kk}$ over the same grid.  To do this,
the Laplacian is first discretized in the usual way.  Letting $f$
denote the right-hand side of~\eqref{eq:backsub}, assuming $I$ has
been discretized such that $(x_i,y_j)$ represents the $(i,j)$th grid
point, and assuming unit distance between adjacent grid points, this
yields the following finite system of linear equations $\forall
(i,j)$:
\begin{align}
  f(x_i, y_j) = & - (U_{kk} + 4) z_{kk}(x_i,y_j) \notag\\
  & + \sum_{\Vert \delta \Vert_1 = 1, \delta \in \mathbb Z^2}
  z_{kk}(x_{i + \delta_1}, y_{j + \delta_2}).
\end{align}
Assuming zero boundary conditions, this system can be solved by
a discrete sine transform.  Since the above expression
can be written as a convolution of $z_{kk}$ with some filter $F$,
this is a deconvolution problem to find $z_{kk}$ given $f$,
and it can be solved by the aforementioned transform, multiply,
inverse-transform method.

\subsection{Reparameterization}

In practice, it was found that a naive parameterization of $B$ and $Q$
caused optimization problems in learning due to ill-conditioning.
Figure~\ref{fig:greens} hints at the reason for this: the amount of
smoothing varies dramatically as the eigenvalues of $(\Bo)^{-1}\Qo$
vary in a very small region around zero.  An exponential change of
coordinates helps to remedy this situation.  Specifically, matrices
$\bBo$ and $\bQo$ are defined such that $\Bo = \exp \bBo$ and
$\Qo = \exp \bQo$, where $\exp$ refers to the matrix exponential. The
learning optimization is then performed in the variables $\bBo$ and $\bQo$.

The loss derivatives with respect to the new variables can be computed
as follows.  Define $\od{L}{\Qo{}}$ as in~\eqref{eq:dLdQ}.  Let
$\bQo = U \Lambda U\tran$ be an eigendecomposition of $\bQo$, defining
$U$ and $\Lambda$.  Then, using known results
for the derivative of the matrix exponential~\cite{najfeld1995derivatives},
it can be shown that
\begin{equation}
  \dod{L}{\bQo{}} =
  U\left(\left(U\tran \dod{L}{\Qo{}}\tran U\right) \odot \Phi\right) U\tran, \label{eq:dLdbQo}
\end{equation}
where $\odot$ is the Hadamard (elementwise) product and $\Phi$ is defined
as follows (defining $\lambda_i := \Lambda_{ii}$):
\begin{equation}
  \Phi_{ij} = 
  \left\{
    \begin{array}{ll}
      (e^{\lambda_i} - e^{\lambda_j}) / (\lambda_i - \lambda_j) & \textrm{if } \lambda_i \neq \lambda_j\\
      e^{\lambda_i} & \textrm{if } \lambda_i = \lambda_j
    \end{array}
    \right..
\end{equation}
For the above to hold, and for $\Qo$ to be positive definite, $\bQo$
must also be symmetric.  This can be enforced by a final transformation,
representing $\bQo$ as the sum of another matrix and its transpose.  It
is then straightforward to apply the chain rule to find the derivative
with respect to that parameterization.  The expression for $\od{L}{\bBo{}}$
is analogous to~\eqref{eq:dLdbQo}.

\subsection{Computational complexity}

Computationally, inference can be decomposed into three components:
performing the Schur decomposition, performing a change of basis,
and solving the scalar PDEs via backsubstitution.  The cost
of performing the Schur decomposition is $O(\No^3)$~\cite{golub2012matrix}.
Let $L$ denote the total number of points in the lattice discretization
(i.e., the number of pixels in the image).
The change of basis $z = V\tran \so$ (and its inverse) consists of
transforming a $\No$-dimensional vector via the square matrix $V$
at each lattice point, at a cost of $O(\No^2 L)$.
The backsubstitution procedure~\eqref{eq:backsub} consists of
computing the right-hand side, which costs $O(\No^2L)$,
and solving the $\No$ scalar PDEs.  Solving each via the
DST costs $O(L\log L)$.  Comptuing $\spr$ costs an additional
$O(\No \Ni L)$. The total computational complexity of
inference is therefore $O(\No^3 + \No^2L + \No L\log L + \No \Ni L)$.
Computing the derivatives for learning requires the same
amount of work, plus an additional $O(L)$ work to 
compute each component of~\eqref{eq:dLdQ} and~\eqref{eq:dLdB}.
The asymptotic complexity is therefore the same as for inference.

\section{Derivation} \label{sec:derivation}

The results of Section~\ref{sec:overview} are now derived.

\subsection{Deriving the reaction-diffusion PDE}

First, the convexity of~\eqref{eq:obj} is easily shown under the
assumption of positive-semidefiniteness of $B$ and $Q$ (by showing
Jensen's inequality holds---proof omitted).  This implies that
any stationary point of~\eqref{eq:obj} must be optimal.
We can find a stationary point by first finding the linear
part of~\eqref{eq:obj} for small variations (i.e., the Fr\'{e}chet
derivative) and then equating this to zero.  Denote by
$J : C^2(\mathbb R^2; \mathbb R^{\Ni+\No}) \rightarrow \mathbb R$
the objective function in~\eqref{eq:obj}, which maps a
twice-differentiable, vector-valued function on $\mathbb R^2$ to a scalar.
Let the notation $\dif f_x$ represent the derivative of a function $f$
at a point $x$, so that 
$\dif J_s : C^2(\mathbb R^2; \mathbb R^{\Ni+\No}) \rightarrow \mathbb R$
represents the derivative of $J$ at $s$.  $\dif J_s$ is identified
as the coefficient of $\epsilon$ in $J(s + \epsilon v)$, where $v$
is an arbitrary variation.  This yields
\begin{equation}
  \dif J_s(v) = 2 \int_I v\tran Q s +
  \sum_{k=1}^2 \dpd{v}{x_k}\tran B \dpd{s}{x_k} \dif x,
\end{equation}
which is verified to be the derivative, as it is linear in $v$.
We now wish to express the term involving $B$ as an inner product with $v$.
To do so, we first rewrite this term (dropping the 2) as
\begin{equation}
  \int_I \sum_{i,j = 1}^{\Ni + \No} \nabla v_i \tran \nabla s_j B_{ij} \dif x.
\end{equation}
We then apply Green's identity
\begin{equation}
  \int_I \psi \Delta \phi + \nabla \psi \tran \nabla \phi \dif x
  = \int_{\partial I} \psi \nabla \phi\tran \hat n \dif S
\end{equation}
for $\phi = v_i$, $\psi = s_j$, use the fact that $v_i = 0$ on $\partial I$,
and regroup terms to obtain
\begin{equation}
  \dif J_s(v) = 2 \int_I v\tran ( Q s - B \Delta s) \dif x.
\end{equation}
Stationarity requires that this be zero on the subspace of feasible
variations, which consists of those variations that do not change the $\si$
components, as these are assumed fixed.  Decomposing $v$ as
$v = \begin{pmatrix} {v^o}\tran & {v^i}\tran \end{pmatrix}\tran$,
we have $v^i = 0$ and
\begin{equation}
  \int_I {v^o}\tran ( \Qo \so + \Qi \si - \Bo \Delta \so - \Bi \Delta \si) \dif x = 0.
\end{equation}
Finally, applying the fundamental lemma of the calculus of variations
yields~\eqref{eq:rdSysPDE}.

\subsection{Solving the PDE system}

The next step is to reduce the solution of~\eqref{eq:rdSysPDE} to
a sequence of scalar PDE subproblems.  Again defining $\spr = \Qi \si - \Bi \Delta \si$,
we left-multiply~\eqref{eq:rdSysPDE} by $(\Bo)^{-1}$ to obtain
\begin{equation}
  \Delta \so - (\Bo)^{-1} \Qo \so = (\Bo)^{-1} \spr.
\end{equation}
We then use the Schur decomposition~\cite{golub2012matrix} to write
$(\Bo)^{-1} \Qo = V U V\tran$, where $V$ is orthonormal.  The assumption that
$B$ and $Q$ are positive-definite implies that $\Bo$ and $\Qo$ are also
positive definite, which implies that 
$(\Bo)^{-1} \Qo$ has a complete set of real, positive eigenvalues
(equal to those of $\sqrt{\Bo} \Qo \sqrt{\Bo}$, which is
positive-definite).  By the properties of the Schur decomposition, $U$
will therefore be upper-triangular, with the positive eigenvalues
of $(\Bo)^{-1} \Qo$ on the diagonal.  Substituting the Schur decompsition
and left-multiplying by $V\tran$ yields
\begin{equation}
  V\tran \Delta \so - UV\tran \so = V\tran (\Bo)^{-1} \spr.
\end{equation}
The next, key step is to observe that the vector Laplacian commutes
with constant linear transformations: i.e., $V\tran \Delta \so = \Delta V\tran
\so$.  This is straightforward to show by expanding the definitions of
matrix multiplication and the vector Laplacian.  This allows us to
perform the change of coordinates $z = V\tran \so$, and to solve for
$z$ instead.  The fact that $U$ is upper-triangular allows us to solve
for $z$ via the backsubstitution algorithm in~\eqref{eq:backsub}.

\subsection{Derivatives}

Inference is regarded as a function mapping \spr{} (as previously
defined) to \so{} by solving~\eqref{eq:rdSysPDE}.  The fact that this
function is well-defined follows from the uniqueness of solutions
of~\eqref{eq:rdSysPDE}, which follows from the assumption that
$B$ and $Q$ are positive definite, and the fact that the scalar
PDEs~\eqref{eq:backsub} have unique solutions~\cite{evans2010partial};
in other words, the linear differential operator $\Bo \Delta - \Qo$
is invertible.  Let $G := (\Bo \Delta - \Qo)^{-1}$.  We now assume a
loss $L$ is defined on the output of $G$, and we wish to find
the derivatives of the loss with respect to the input of $G$
(i.e., the backpropagation derivative) as well as the
derivatives with respect to the parameters $\Bo$ and $\Qo$.

First, we assume that the derivative of $L$ is provided in the form
of a function $\od{L}{{\so}} : I \rightarrow \mathbb R^\No$:
\begin{equation}
  (\dif L_{\so{}}) v = \left< \od{L}{{\so}}, v \right>,
\end{equation}
where $v$ is an arbitrary variation.  Intuitively, $\od{L}{{\so}}$
represents the differential change in $L$ due to a variation of $\so$
at the point given by its input.  We would like to obtain the
derivative of $L \circ G$ in the same form.  By the chain rule and the
definition of the adjoint,
\begin{align}
  (\dif (L \circ G)_{\spr{}}) v
  &= \left< \od{L}{\so{}}, (\dif G_{\spr{}}) v \right> \notag\\
  & = \left< (\dif G_{\spr{}})^* \od{L}{\so{}}, v \right>.
\end{align}
Since $G$ is linear in $\spr{}$, $\dif G_{\spr{}} = G$.  Furthermore,
$G$ is self-adjoint; this follows from the fact that $\Bo \Delta -
\Qo$ is self-adjoint, which in turn can be shown given that $\Bo$ and
$\Qo$ are self-adjoint (by the assumption of positive-definiteness)
and $\Bo$ commutes with $\Delta$.  This implies $\od{L}{\spr{}} =
G\od{L}{\so{}}$, which is equivalent to~\eqref{eq:dLdsp}.

To obtain the parameter derivatives, we directly consider the effect
of adding a small perturbation to each parameter.  We first define the
unperturbed solution $\so{}$ as the solution to $(\Bo \Delta - \Qo)
\so = \spr$, given the input $\spr$.  We then define the perturbed
solution $\tso$ as the solution to $((\Bo +\epsilon V) \Delta - \Qo)
\tso = \spr$, where $\epsilon V$ is a variation of $\Bo$.  We then
find the following expansion of $\tso$ in $\epsilon$.  In the following,
the notation $G_{\Bo}$ is used to refer to $G$ evaluated with the
parameter $\Bo$.
\begin{align}
  (\Bo \Delta - \Qo) \tso & = \spr - \epsilon V \Delta \tso \notag\\
  \tso & = G_{\Bo} ( \spr - \epsilon V \Delta \tso) \notag\\
  & = \so - \epsilon G_{\Bo} V \Delta \tso \notag\\
  & = \so - \epsilon G_{\Bo} V \Delta (\so - \epsilon G_{\Bo} V \Delta \tso)\notag\\
  & = \so - \epsilon G_{\Bo} V \Delta \so + O(\epsilon^2)
\end{align}
Note that the preceding two lines are obtained by recursive expansion.
This implies that $(\dif G_{\Bo})V = -G_{\Bo} V \Delta \so$, 
again abusing notation so that $\dif G_{\Bo}$ refers to the derivative of $G$
as a function of the parameter $\Bo$, and evaluated at the point $\Bo$.
The chain rule and adjoint property are applied again to obtain
\begin{align}
  (\dif (L \circ G)_{\Bo}) V & = \left< -G_{\Bo} \dod{L}{\so{}}, V \Delta \so \right>\\
  & = \left< -\dod{L}{\spr{}}, V \Delta \so \right>.
\end{align}
The previous arguments apply even in the case that $\Bo$ is a function
of $I$ (i.e., depends on $x$).  If $\Bo$ is constrained to be constant
on $I$, however, the variations $V$ must also be constant, and we can
write a basis for these in the form $\mathbf 1_i \mathbf 1_j\tran$,
for $(i,j) \in \{1,\dots,\No\}^2$.  We then define $\od{L}{\Bo_{ij}{}}
= (\dif (L \circ G)_{\Bo}) \mathbf 1_i \mathbf 1_j\tran$.  Evaluating
this using the expression above then yields~\eqref{eq:dLdB}.
Repeating the argument above {\em mutatis mutandis} for \Qo{}
yields~\eqref{eq:dLdQ}.

\section{Related work}

The method proposed here is comparable to recent work on joint
training of CNNs and CRFs.  In particular,~\cite{zheng2015conditional}
and~\cite{schwing2015fully} both propose
backpropagation-through-inference techniques in order to jointly train
fully-connected CRFs (FC-CRFs)~\cite{fccrf2011} and CNNs.  Inference
consists of performing a fixed number of message-passing iterations on
the pseudo-marginals of the mean-field approximation.  These
message-passing steps are then regarded as a formal composition of
functions, to which backpropagation is applied.  The FC-CRF model is
probably more expressive than \method{}, in that it features non-local
connections.  However, this comes at the expense of having to resort
to approximate inference.  The asymptotic complexity of performing a
single message-passing round in the FC-CRF is comparable to that of
completing the entire exact inference procedure in \method{}.  The
method of~\cite{ranftl2014deep} is also notable for jointly training a
CNN with inference based on a global convex energy model, but inference 
and learning in this model rely on general optimization techniques that
are costly when dealing with image-sized optimization problems.

Gaussian random field models have previously been applied to
the task of semantic segmentation~\cite{tappen2008logistic}
and related tasks~\cite{tappen2007utilizing,jancsary2012regression};
however, exact inference in these methods scales at least quadratically
in the number of pixels, making approximate inference necessary again.

Recent work~\cite{chen2015learning} learned reaction-diffusion processes
for image restoration; however, this work relied on simulating the process
in time to do inference and learning, resulting in a method very similar
to the aforementioned backpropagation-through-inference techniques.

Solving the PDE~\eqref{eq:rdSysPDE} is related to Wiener filtering. It
is known that certain {\em scalar-valued} Gaussian MRFs may be solved
via Wiener filtering~\cite{szeliski2010computer}; to our knowledge,
the inference procedure in this work is novel in that it essentially
reduces inference in a {\em vector-valued} Gaussian MRF to repeated
Wiener filtering.

\section{Experiments} \label{sec:experiments}

\method{} was implemented in Caffe~\cite{jia2014caffe} and compared to
several other methods on two datasets: the KITTI road segmentation
dataset~\cite{Fritsch2013ITSC} (a binary classification problem) and
the Stanford background dataset~\cite{gould2009decomposing} (an 8-way
classification problem).  The experiments were conducted with the
following primary goals in mind: (1) to directly compare \method{} to
an global energy method based on ``backpropagation-through-approximate-inference;''
(2) to investigate whether composing multiple \method{}
layers is feasible and/or beneficial; and (3) to see whether
integration of \method{} with deep CNNs might provide any benefit.

All of the parameters (CRF, \method{}, and convolution) in all of the
experiments were trained jointly via backpropagation, using the
AdaGrad algorithm~\cite{duchi2011adaptive}.  Each experiment
consisted of training all parameters from scratch---i.e., using random
initialization and no other training data besides---and all
experiments were run until the loss converged (using a
stepwise-annealed learning rate).  The softmax loss
was used for all experiments. \method{} was
compared to the previously discussed method
of~\cite{zheng2015conditional}, referred to here as CRF-RNN, which
performs backpropagation through approximate inference for a FC-CRF.
The default number of mean-field iterations (10) was used for this
method.  The authors' public implementation was used.  The method
of~\cite{googlenet}, referred to here as GoogleNet, was also
evaluated.  The public BVLC implementation was used, slightly modified
for semantic segmentation by transforming it into a
fully-convolutional model~\cite{long2015fully}.  A single random split
of each dataset was chosen, with 80\% of the data reserved for training
and the remaining held-out.  Each experiment reports the results of
evaluation on this held-out set.

\begin{figure}
  \centering
  \begin{tabular}{cc}
    \multirow{2}{*}[0.75in]{
      \subfloat[Layered \method{}]{
        \includegraphics[height=2.0in]{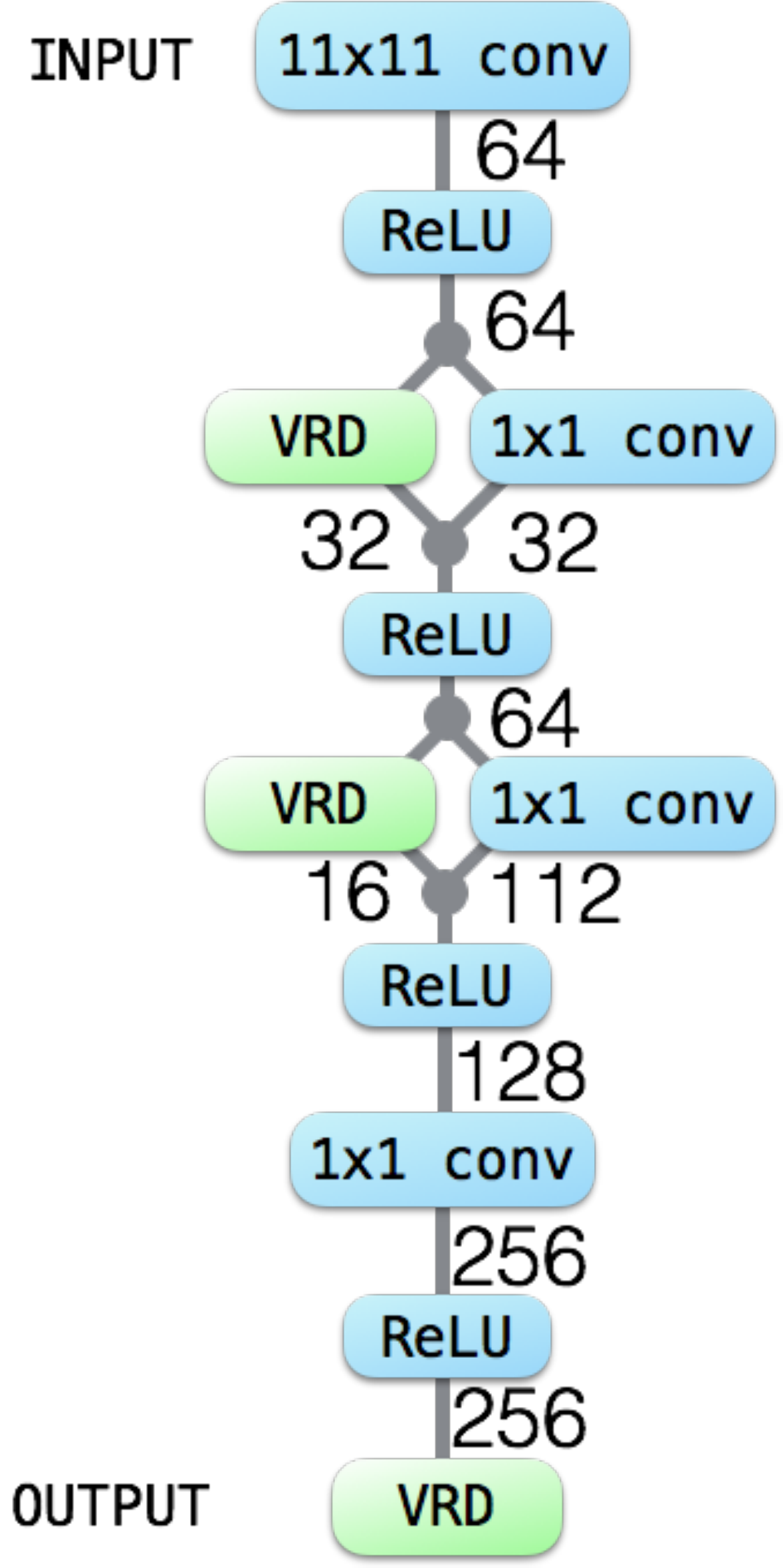}
      }
    } &
    \subfloat[Shallow nets]{
      \includegraphics[height=0.65in]{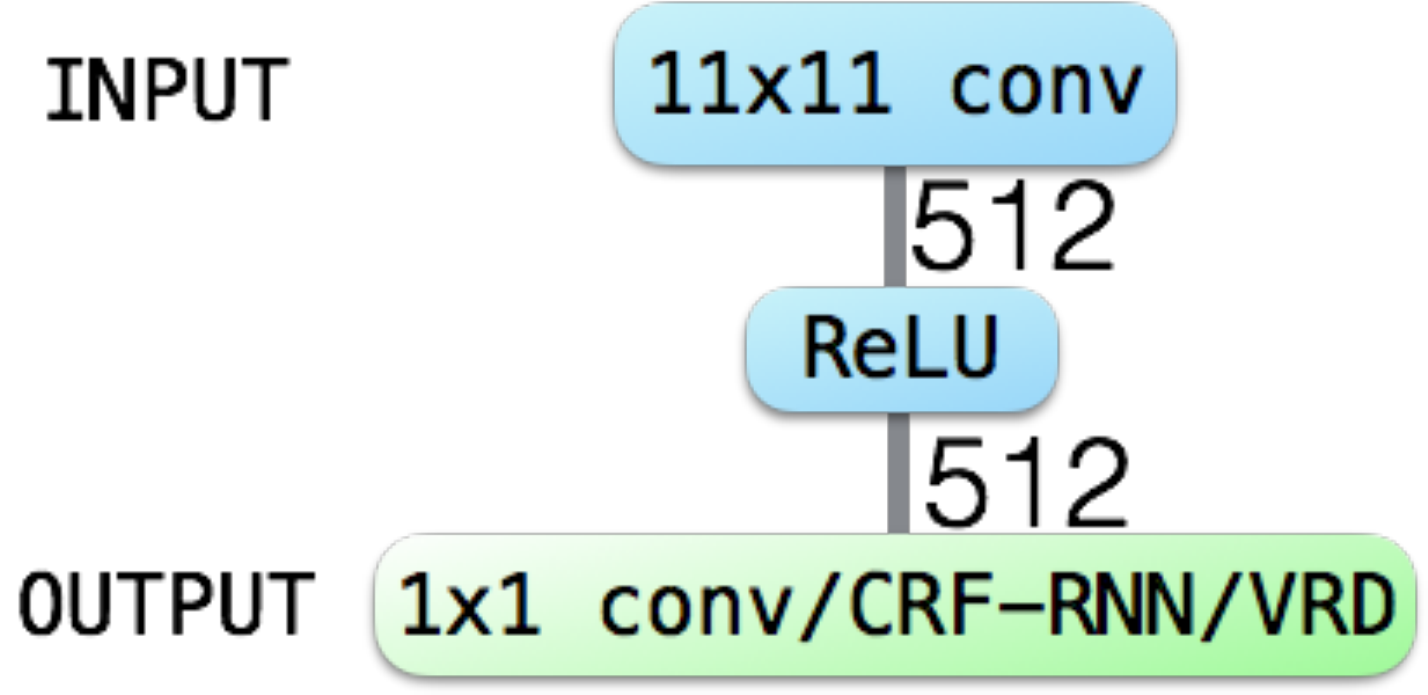}
    }\\
    & \subfloat[GoogleNet + CRF/\method{}]{
      \includegraphics[height=1.0in]{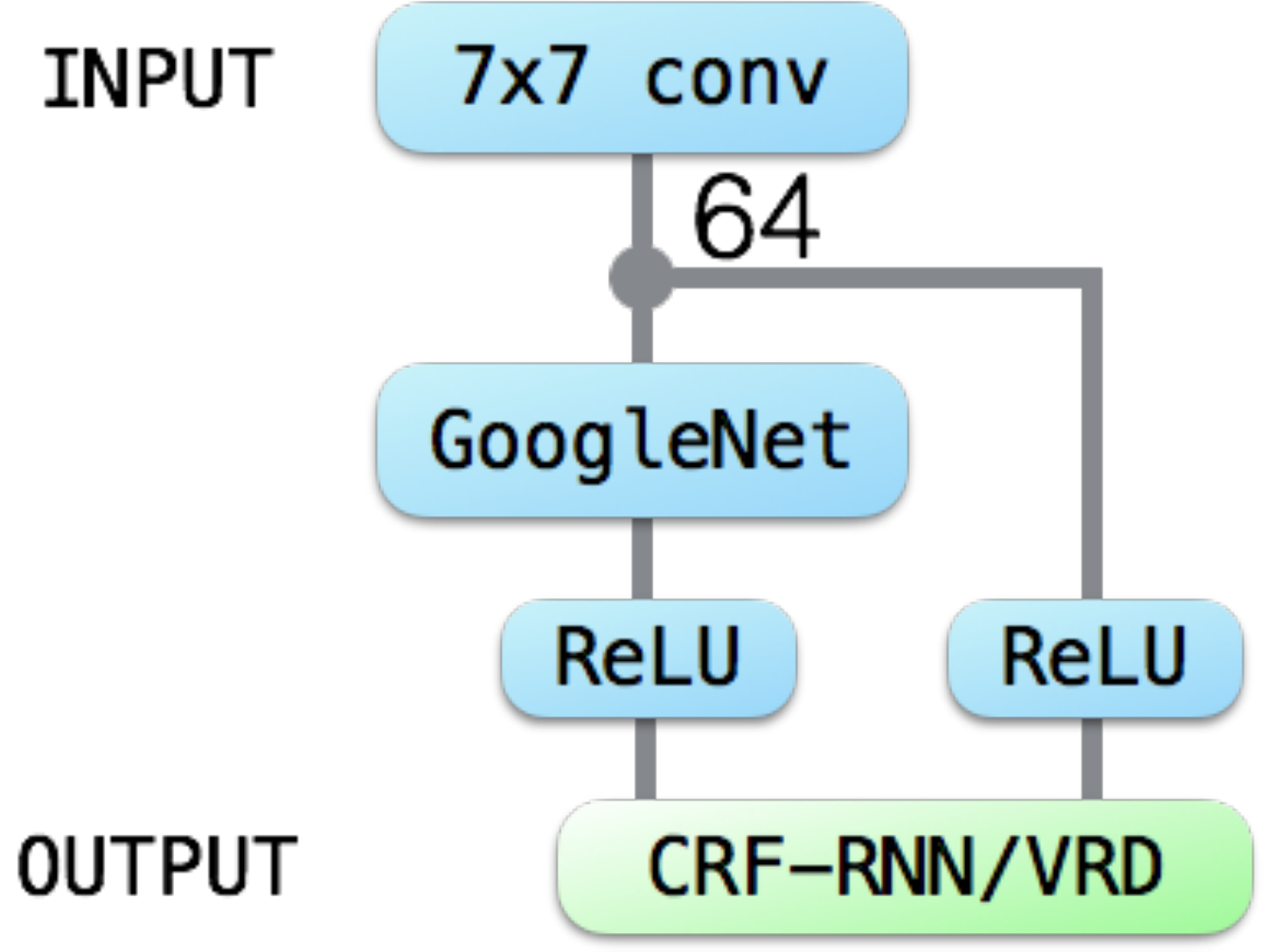}
    }
  \end{tabular}
  \caption{Architectures used in experiments. Numbers under layers indicate
    number of output channels produced.}
  \label{fig:arch}
\end{figure}

Each experiment involved training a CNN with one of three general
architectures, which are illustrated in Fig.~\ref{fig:arch}.  Each
type of architecture was designed primarily with the goal of testing
one of the questions mentioned above.  Experiments with shallow
architectures were designed to directly compare the merits of
different global energy models.  The {\em Layered VRD} experiments
were intended to test whether layering \method{} layers is feasible and
provides any benefit (the {\em layered baseline} experiment is
identical to Layered \method{}, but exchanging the \method{} layers with
1x1 convolutions).  The {\em GoogleNet} experiments were
intended to test whether \method{} is useful in the context of
joint training with a deep CNN.

\begin{figure}
  \centering
  \includegraphics[width=3.25in]{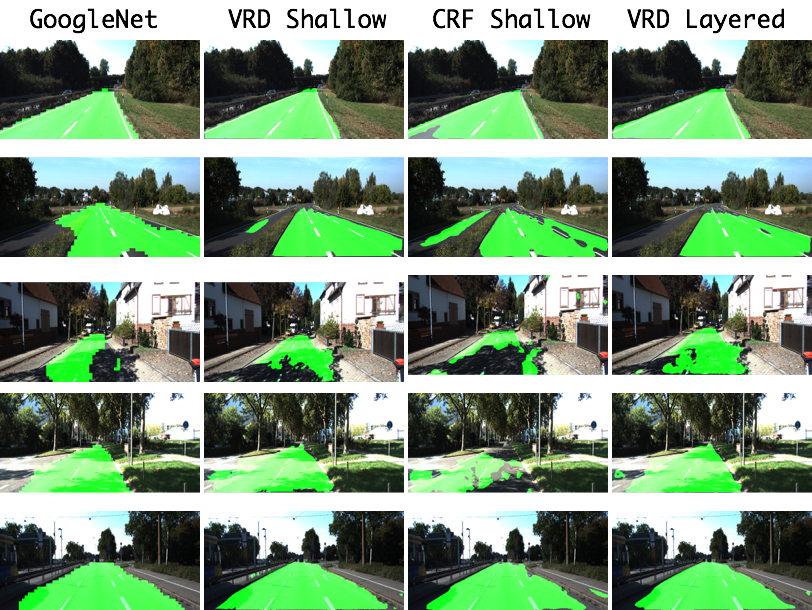}
  \caption{Qualitative results for KITTI dataset.}
  \label{fig:kittiQual}
\end{figure}

\begin{figure}
  \centering
  \includegraphics[width=3.25in]{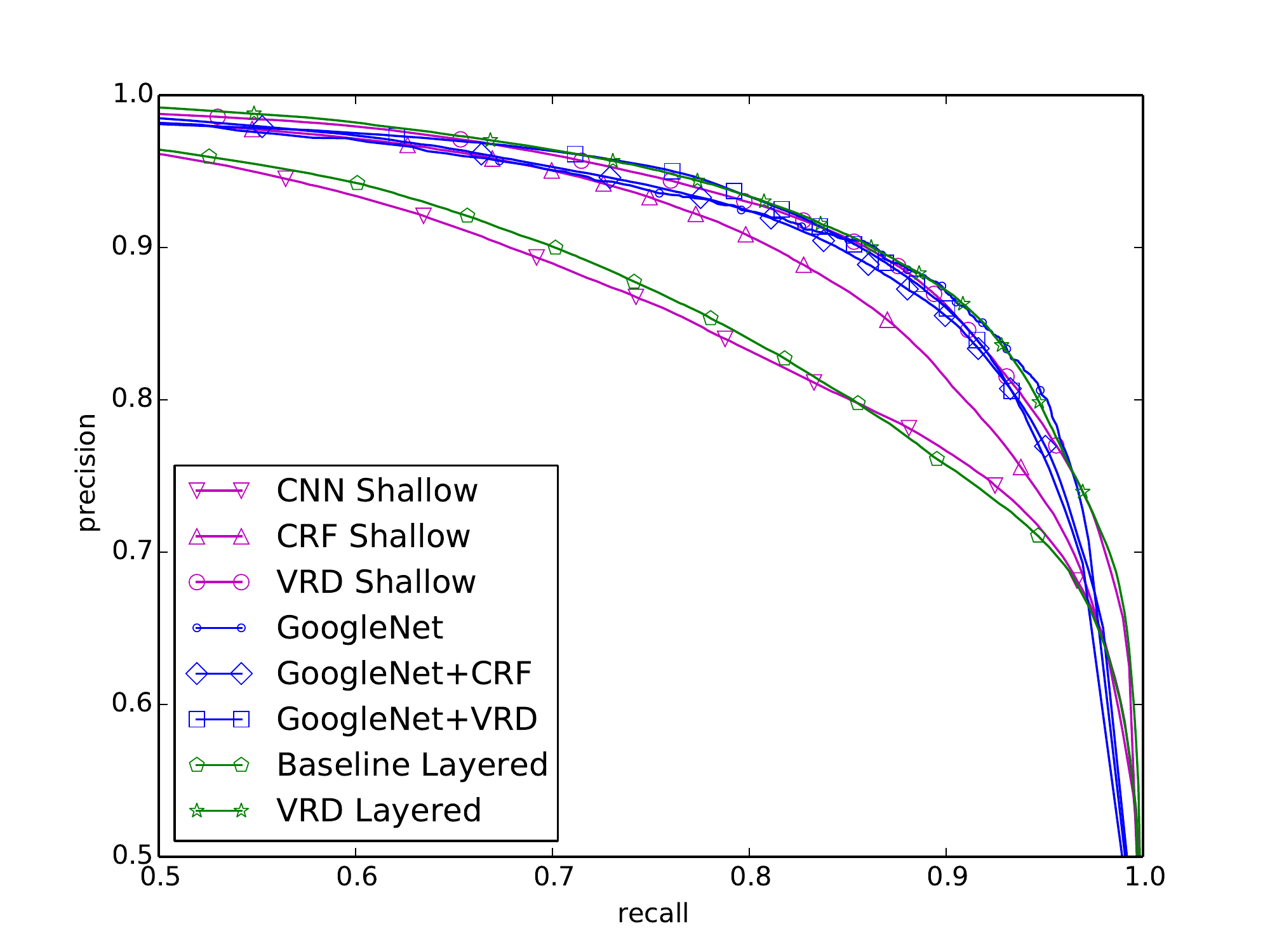}
  \caption{Precision-recall curves for KITTI experiments.}
  \label{fig:prc}
\end{figure}

Precision-recall curves for the KITTI experiments are shown in
Fig.~\ref{fig:prc}, and qualitative results are shown in
Fig.~\ref{fig:kittiQual}.  Table~\ref{tab:results} lists evaluation
results for both datasets: for KITTI, maximum F1 and AP are reported
in the birds-eye view (as suggested in~\cite{Fritsch2013ITSC}); and
for Stanford Background (abbreviated SBG), pixel-level accuracies are
reported.  Comparing \method{} and CRF-RNN, it was consistently
observed that better results were obtained using \method{}.  The most
significant benefits were observed with shallow architectures,
although a significant benefit was also observed in the experiments
with GoogleNet on SBG.  Layering \method{} proved to be both practical
and beneficial.  On the KITTI dataset, this method seemed to produce
the best overall results, despite having far fewer parameters and a
far simpler architecture than GoogleNet.  However, this architecture
had too few parameters to fit the SBG data well.  Finally, joint
training of \method{} with GoogleNet was not beneficial for the KITTI
dataset; however, a large benefit was seen on the more difficult SBG
dataset.

\begin{table}
  \centering
  \begin{footnotesize}
  \begin{tabular}{llll}
    & \multicolumn{2}{c}{KITTI} & SBG\\
    Method & Max F1 & AP & Acc.\\ \hline\hline
    Shallow CNN & 82.84 & 88.29 & 56.1\\
    Shallow CRF-RNN & 86.14 & 90.64 & 59.3\\
    Shallow \method{} & 88.22 & 91.76 & 62.7\\
    GoogleNet & {\bf 88.62} & 91.25 & 65.0\\
    GoogleNet + CRF-RNN & 87.72  & 91.49 & 68.2\\
    GoogleNet + \method{} & 88.07 & 91.38 & {\bf 70.7}\\
    Layered baseline & 82.55 & 88.56 & 55.2 \\
    Layered \method{} & 88.58 & {\bf 92.14 } & 61.5
  \end{tabular}
  \end{footnotesize}
  \caption{Experimental results}
  \label{tab:results}
\end{table}

To give a general idea of the computational efficiency of \method{},
for an internal \method{} layer with $\Ni = 64$, $\No = 32$, and an image
size of 511x255 $(L = 130305)$, forward-pass inference took $408$ ms, and
computing derivatives in the backwards pass took $760$ ms. These 
timings are for a CPU implementation, although $\spr$ was computed
via GPU convolutions.

\section{Conclusions}

A global energy model for semantic segmentation featuring very
efficient exact inference was proposed.  Inference in \method{} for a
problem with $\No$ output labels reduces to a sequence of $\No$
convolutions, which can be implemented efficiently via the FFT, and
backpropagation and parameter derivatives for learning can be computed
just as efficiently, making it an attractive choice for joint training with CNNs.

Analysis revealed how \method{} can be thought of as a relaxation of
other global energy methods.  Despite this, experiments demonstrated
superior performance of \method{} 
compared to a more complex FC-CRF-based model in the
context of joint training with a CNN.  This suggests that, at least in
the tested scenarios, the benefits of exact inference may outweigh
those of having a more expressive or sophisticated model.  The
experiments also demonstrated the feasibility of composing multiple
\method{} layers, and this yielded promising results.

In the short term, more work needs to be done to devise and test CNN
architectures that are able to leverage the ability of \method{} to
efficiently produce and backpropagate through exact global inferences.
In the longer term, it is hoped that the insights developed here will
lead to a better general understanding of how best to integrate CNNs
with global energy models.

\bibliography{vrd}

\begin{thebibliography}{10}

\bibitem{brandt1982guide}
Achi Brandt.
\newblock Guide to multigrid development.
\newblock In {\em Multigrid methods}, pages 220--312. Springer, 1982.

\bibitem{chen14semantic}
Liang-Chieh Chen, George Papandreou, Iasonas Kokkinos, Kevin Murphy, and Alan~L
  Yuille.
\newblock Semantic image segmentation with deep convolutional nets and fully
  connected crfs.
\newblock In {\em ICLR}, 2015.

\bibitem{chen2015learning}
Yunjin Chen, Wei Yu, and Thomas Pock.
\newblock On learning optimized reaction diffusion processes for effective
  image restoration.
\newblock In {\em Proceedings of the IEEE Conference on Computer Vision and
  Pattern Recognition}, pages 5261--5269, 2015.

\bibitem{duchi2011adaptive}
John Duchi, Elad Hazan, and Yoram Singer.
\newblock Adaptive subgradient methods for online learning and stochastic
  optimization.
\newblock {\em The Journal of Machine Learning Research}, 12:2121--2159, 2011.

\bibitem{evans2010partial}
Lawrence~C Evans.
\newblock {\em Partial differential equations}.
\newblock American Mathematical Society, 2010.

\bibitem{Fritsch2013ITSC}
Jannik Fritsch, Tobias Kuehnl, and Andreas Geiger.
\newblock A new performance measure and evaluation benchmark for road detection
  algorithms.
\newblock In {\em International Conference on Intelligent Transportation
  Systems (ITSC)}, 2013.

\bibitem{golub2012matrix}
Gene~H Golub and Charles~F Van~Loan.
\newblock {\em Matrix computations}, volume~3.
\newblock JHU Press, 2012.

\bibitem{gould2009decomposing}
Stephen Gould, Richard Fulton, and Daphne Koller.
\newblock Decomposing a scene into geometric and semantically consistent
  regions.
\newblock In {\em Computer Vision, 2009 IEEE 12th International Conference on},
  pages 1--8. IEEE, 2009.

\bibitem{jancsary2012regression}
Jeremy Jancsary, Sebastian Nowozin, Toby Sharp, and Carsten Rother.
\newblock Regression tree fields—an efficient, non-parametric approach to
  image labeling problems.
\newblock In {\em Computer Vision and Pattern Recognition (CVPR), 2012 IEEE
  Conference on}, pages 2376--2383. IEEE, 2012.

\bibitem{jia2014caffe}
Yangqing Jia, Evan Shelhamer, Jeff Donahue, Sergey Karayev, Jonathan Long, Ross
  Girshick, Sergio Guadarrama, and Trevor Darrell.
\newblock Caffe: Convolutional architecture for fast feature embedding.
\newblock {\em arXiv preprint arXiv:1408.5093}, 2014.

\bibitem{fccrf2011}
Philipp Kr\"{a}henb\"{u}hl and Vladlen Koltun.
\newblock Efficient inference in fully connected crfs with gaussian edge
  potentials.
\newblock In J.~Shawe-Taylor, R.S. Zemel, P.L. Bartlett, F.~Pereira, and K.Q.
  Weinberger, editors, {\em Advances in Neural Information Processing Systems
  24}, pages 109--117. Curran Associates, Inc., 2011.

\bibitem{lin2015Piecewise}
Guosheng Lin, Chunhua Shen, Ian~D. Reid, and Anton van~den Hengel.
\newblock Efficient piecewise training of deep structured models for semantic
  segmentation.
\newblock {\em CoRR}, abs/1504.01013, 2015.

\bibitem{long2015fully}
Jonathan Long, Evan Shelhamer, and Trevor Darrell.
\newblock Fully convolutional networks for semantic segmentation.
\newblock In {\em Proceedings of the IEEE Conference on Computer Vision and
  Pattern Recognition}, pages 3431--3440, 2015.

\bibitem{najfeld1995derivatives}
Igor Najfeld and Timothy~F Havel.
\newblock Derivatives of the matrix exponential and their computation.
\newblock {\em Advances in Applied Mathematics}, 16(3):321--375, 1995.

\bibitem{ranftl2014deep}
Ren{\'e} Ranftl and Thomas Pock.
\newblock A deep variational model for image segmentation.
\newblock In {\em Pattern Recognition}, pages 107--118. Springer, 2014.

\bibitem{schwing2015fully}
Alexander~G Schwing and Raquel Urtasun.
\newblock Fully connected deep structured networks.
\newblock {\em arXiv preprint arXiv:1503.02351}, 2015.

\bibitem{googlenet}
Christian Szegedy, Wei Liu, Yangqing Jia, Pierre Sermanet, Scott Reed, Dragomir
  Anguelov, Dumitru Erhan, Vincent Vanhoucke, and Andrew Rabinovich.
\newblock Going deeper with convolutions.
\newblock {\em CoRR}, abs/1409.4842, 2014.

\bibitem{szeliski2010computer}
Richard Szeliski.
\newblock {\em Computer vision: algorithms and applications}.
\newblock Springer Science \& Business Media, 2010.

\bibitem{tappen2007utilizing}
Marshall~F Tappen.
\newblock Utilizing variational optimization to learn markov random fields.
\newblock In {\em Computer Vision and Pattern Recognition, 2007. CVPR'07. IEEE
  Conference on}, pages 1--8. IEEE, 2007.

\bibitem{tappen2008logistic}
Marshall~F Tappen, Kegan~GG Samuel, Craig~V Dean, and David~M Lyle.
\newblock The logistic random field—a convenient graphical model for learning
  parameters for mrf-based labeling.
\newblock In {\em Computer Vision and Pattern Recognition, 2008. CVPR 2008.
  IEEE Conference on}, pages 1--8. IEEE, 2008.

\bibitem{zheng2015conditional}
Shuai Zheng, Sadeep Jayasumana, Bernardino Romera-Paredes, Vibhav Vineet,
  Zhizhong Su, Dalong Du, Chang Huang, and Philip~HS Torr.
\newblock Conditional random fields as recurrent neural networks.
\newblock In {\em Proceedings of the IEEE International Conference on Computer
  Vision}, pages 1529--1537, 2015.

\end{thebibliography}
\bibliographystyle{plain}

\end{document}